# Video Swin Transformers for Egocentric Video Understanding @ Ego4D Challenges 2022

Mar´ıa Escobar*
Universidad de Los Andes
Bogotá, Colombia
mc.escobar11@uniandes.edu.co

Laura Daza*
Universidad de Los Andes
Bogotá, Colombia
la.daza10@uniandes.edu.co

Cristina González
Universidad de Los Andes
Bogotá, Colombia
ci.gonzalez10@uniandes.edu.co

Pablo Arbeláez
Universidad de Los Andes
Bogotá, Colombia
pa.arbelaez@uniandes.edu.co

Jordi Pont-Tuset
Google Research
Zürich, Switzerland
jponttuset@google.com

## 1. Video Swin Transformer

Transformers are self-attention architectures that have recently demonstrated superior performances to convolutional neural networks in visual recognition tasks. In particular, Swin Transformers [9] present a shifted window partition mechanism that adds connections between adjacent windows and counteracts the limitation of local operations. This architecture achieves state-of-the-art results for the task of image classification and reduces the computational cost of obtaining self-attention in images by restricting the calculations to non-overlapping windows.

Swin Transformers are extended to the video domain by changing the 2D windows with 3D sub-volumes and eliminating the resolution reductions in the temporal dimension [10].

### 1.1. Deformable Swin Transformer

Even with the decreased computational cost of using shifted windows, the extension of said windows to the temporal dimension for Video Swin Transformers results in greater model complexity. To mitigate this issue, we propose to replace the self-attention within the 3D patches with deformable attention mechanisms. With this strategy, each input position to only $N$ points instead of the whole patch.

## 2. Egocentric video understanding

We participated in the Ego4D challenges for Object State Change Classification [3] and Point-of-No-Return temporal localization [4]. The base code and pretrained models are available at https://github.com/BCV-Uniandes/PNR_OSCC.

### 2.1. Object State Change Classification

#### 2.1.1 Experimental Setting

*Dataset:* for our experiments we use the Ego4D [6] Hands and Objects benchmark. We focus on the object state change classification task, where the goal is to identify if an object changed its state in 8-seconds clips. The dataset for this challenge is composed of 41$K$ clips for training, 28$K$ for validation, and 28$K$ for testing.

*Training details:* we follow the Video Swin Transformer [10] implementation and train our model using AdamW optimizer with learning rate $3e^{-4}$ for 30 epochs. All the transformer attention modules are trained from scratch, while the Multi-Layer Perceptrons in the transformer blocks are initialized from weights pretrained on Kinetics-600 [1, 8]. For the deformable attention, we empirically found that setting $N = 4$ yields good accuracy while preserving a low computational complexity.

#### 2.1.2 Results

In table 2 we evaluate our Deformable Swin Transformer in the Ego4D Hands and Objects benchmark and compare it with the challenge baseline: the state-of-the-art I3D with ResNet-50 backbone [2]. We also compare with the top performing transformer-based models for video classification, namely MViT [5] and Video Swin Transformer. Both methods were trained following the original configuration.

Our results show that MViT has a performance on par with the convolutional neural network I3D. The Video Swin Transformer outperforms the previous method by 0.8% accuracy. Finally, our Deformable Swin Transformers obtains a further improvement of 0.3%, reaching a 69.8 accuracy in the validation set and 67.7 in the test set.

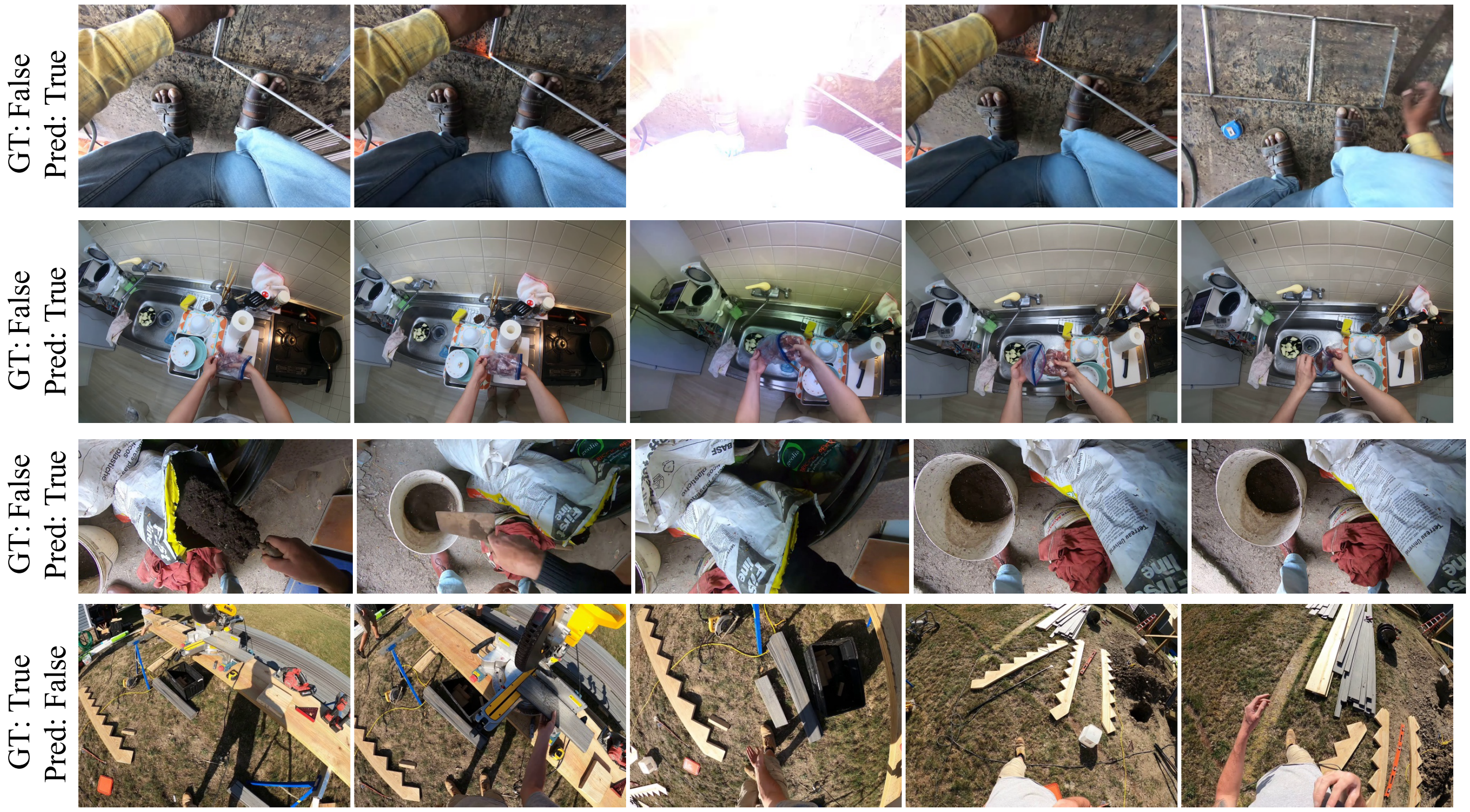

Figure 1. **Failure Cases.** We display some examples of incorrect predictions obtained by our model. The frames are uniformly sampled throughout the clip.

| Method | Validation Acc |
|---|---|
| I3D ResNet-50 [2] | 68.7 |
| MViT [5] | 68.7 |
| Video Swin Transformer [10] | 69.5 |
| Ours | 69.8 |

Table 1. **Comparison with the state-of-the-art.** We compare the performance of our proposed approach with various state-of-the-art methods for video classification. The results show that our Deformable Swin Transformer outperforms the Ego4D baseline (1st row) and the top-performing transformer models for video classification.

In figure 6 we display several examples of failure cases of our model. The first row shows two pieces of metal welded and is classified as a positive case of state change. However, the clip is labeled as a negative case. Rows 2 and 3 present objects changing containers, *i.e.* a piece of meat taken out of a bag and dirt being poured into a bucket. Both examples show interactions with the objects, but their states are not altered. The fourth row shows a piece of wood taken from a table and thrown into the ground, which is classified as a negative state change.

## 2.2. Point-of-No-Return Temporal Localization

### 2.2.1 Baseline

We performed a frequency analysis of the normalized Point-of-No-Return (PNR) across the training and validation clips for our baseline approach. To calculate the normalized PNR, we divided each PNR frame by the entire duration of the corresponding clip. Figures 2 and 3 show that there is a tendency toward having the PNR at the 0.45 fraction of the duration of the video. Thus, our baseline approach to identify the PNR is to multiply the duration in frames of each test video by 0.45.

### 2.2.2 Experimental Setting

*Dataset:* We use the data provided in the PNR temporal localization task of the Ego4D [6] Hands and Objects benchmark. Following the instructions of the authors [6] we used solely the clips that included a state change for the training and validation of our approach. The dataset for this challenge is composed of $20K$ clips for training, $13K$ for validation, and $28K$ for testing.

*Training details:* we follow the Video Swin Transformer [10] implementation and train our model from scratch using AdamW optimizer with learning rate $3e^{-4}$ for 20 epochs. For the PNR task it is crucial to consider the frame sampling strategy since it defines the level of preci-sion in which the PNR frame can be retrieved. We perform a uniform sampling strategy with a fixed input size of 16 frames. Once this sampling is done, we calculate the standard sampling rate, defined as the duration of the video divided by the input size, and use this rate to locate the PNR in the input sequence. For validation and testing, we use the standard sampling rate to decode the location of the PNR frame.

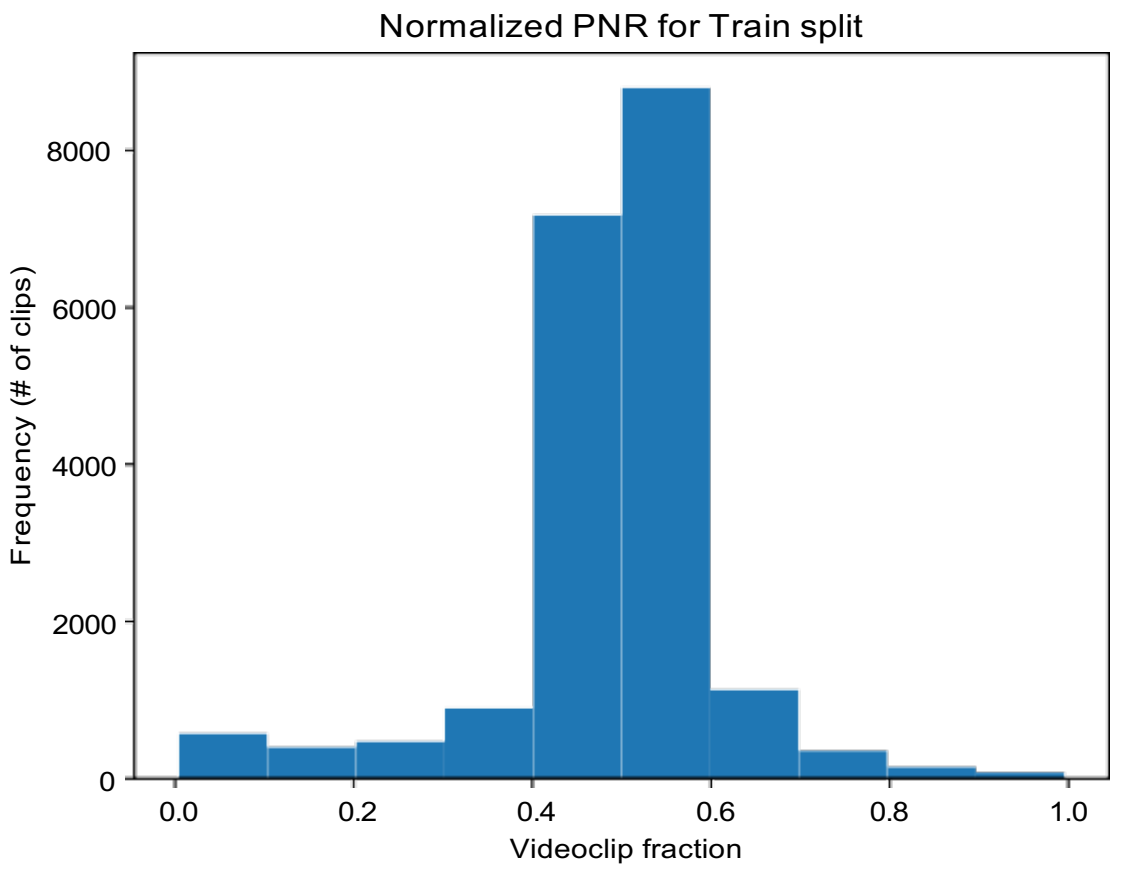

Figure 2. Distribution of normalized PNR across clips for the training split. Most PNR are accumulated towards the 0.45 fraction of the video duration.

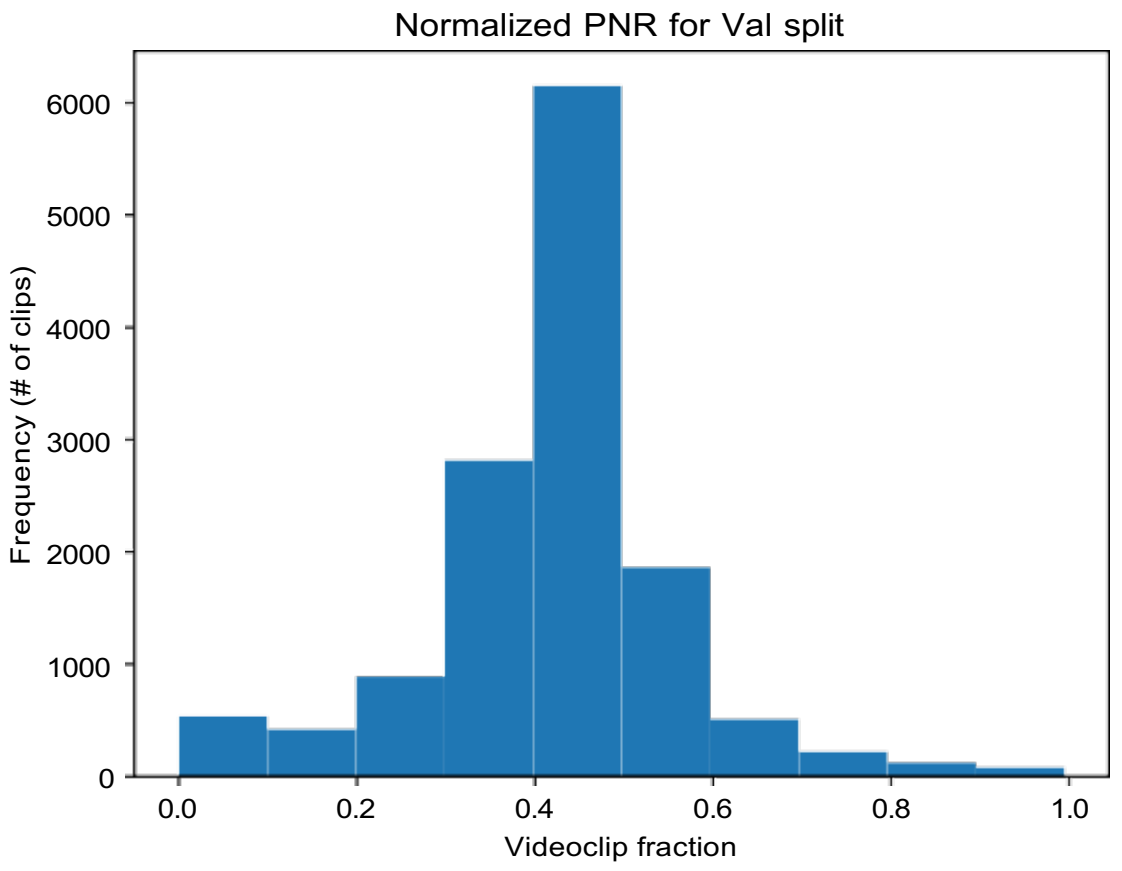

Figure 3. Distribution of normalized PNR across clips for the validation split. Most PNR are accumulated towards the 0.45 fraction of the video duration.

#### 2.2.3 Results

In table 2 we evaluate our baseline and Video Swin Transformer approaches for the PNR temporal localization task and compare them with two baselines provided by the challenge: always selecting the center frame and a Bi-directional LSTM [7]. The results from the test split were obtained through the official evaluation server of the challenge.

| Method | Validation error | Test error |
|---|---|---|
| Always Center Frame [6] | 1.03 | 1.06 |
| Bi-directional LSTM [2] | 0.79 | 0.76 |
| Baseline | 0.63 | 0.67 |
| **Video Swin Transformer [10]** | **0.61** | **0.66** |

Table 2. **Comparison with the state-of-the-art.** We compare the performance of our proposed baseline and final approaches with the baseline methods designed for the challenge. The results show that our Video Swin Transformer outperforms both baselines designed proposed by the challenge. The metric is the absolute temporal localization error (seconds).

Table 2 shows that our proposed baseline of always selecting the frame corresponding to 0.45 times the video duration achieves better results than the Always Center Frame baseline proposed by the challenge. Additionally, our baseline also obtains better performance than the CNN-based method. These results suggest that there is a unintended bias in the selection of the PNR frame for all the splits of the dataset. Furthermore, our Video Swin Transformer implementation achieves a higher performance than all of the proposed baselines, obtaining a final test error of 0.66 seconds.

In figure 5 we display some examples of clips for which our Video Swin Transformer prediction was more accurate than our proposed baseline. The first row shows that Video Swin Transformer localizes the PNR perfectly, while the baseline prediction is not accurate enough. The second row shows an example in which neither the baseline prediction nor the prediction of our final method can locate the PNR correctly. However, the annotated clip does not change significantly throughout time, and there is no clear interaction with the objects.

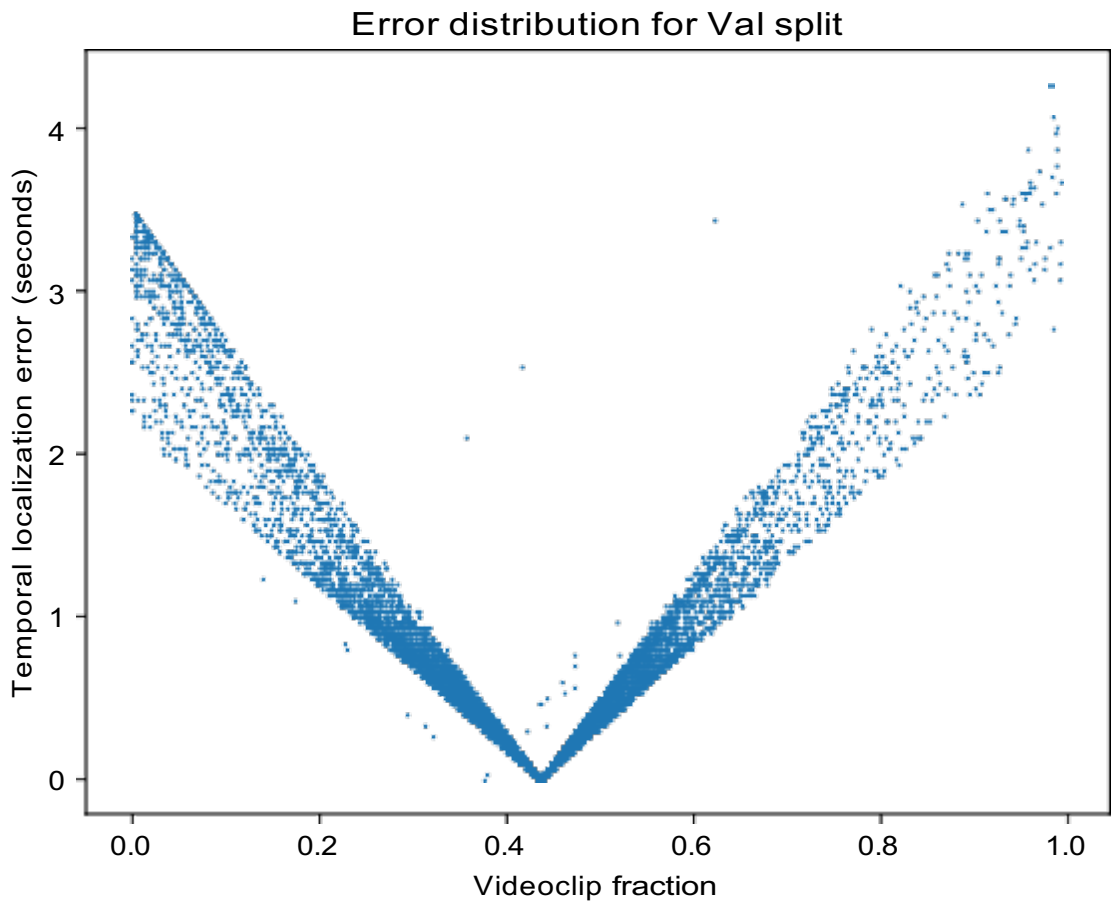

Figure 4. Error distribution for the predictions of our Video Swin Transformer in the validation split.

Figure 6 shows failure cases of our final Video Swin Transformer approach. For both examples, our method predicts that the PNR is near the middle of the clip. However, in both cases, the ground-truth PNR frame is annotated at the end of the clip, with only four frames left, so it is not possible to identify if it correctly corresponds to a state change of the objects.

Finally, figure 4 shows the error distribution for the predictions of our Video Swin Transformer in the validation split. The low error values in the middle show a tendency to predict values close to the 0.45 fraction of the clip length. The clips with the highest errors correspond to those with ground-truth PNR frames at unusual locations, such as in the beginning or the end of a video.

## 3. Conclusion

We implemented Video Swin Transformer as a base architecture for the tasks of Point-of-No-Return temporal localization and Object State Change Classification. Our method achieved competitive performance on both challenges.

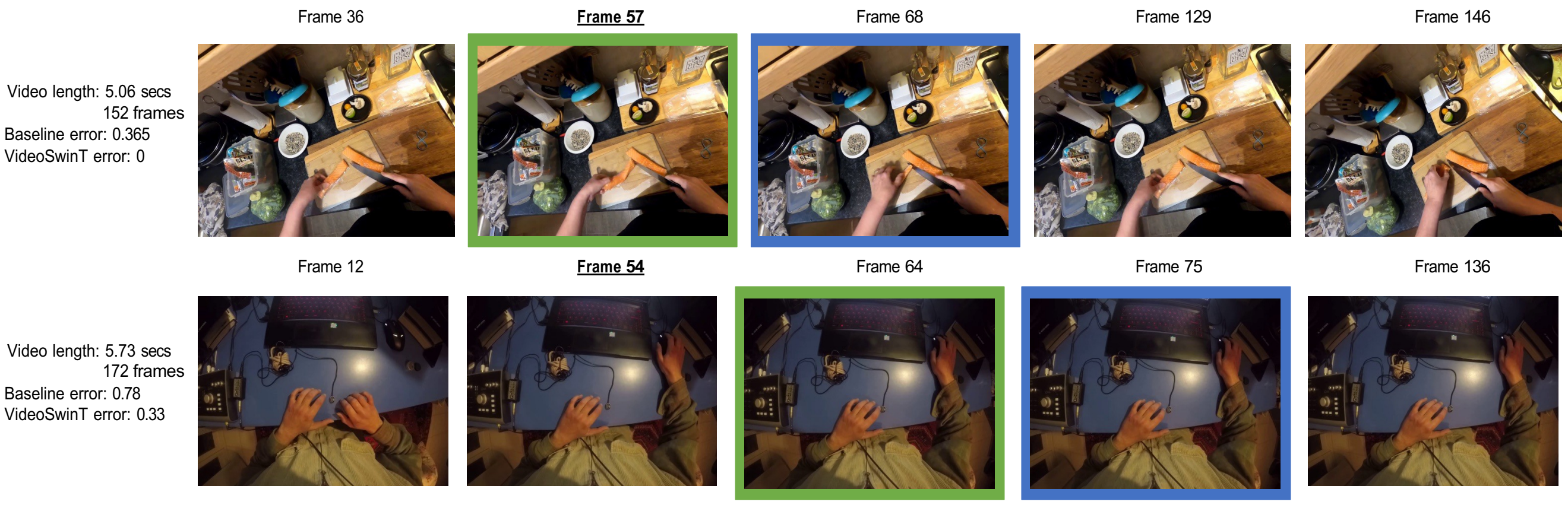

Figure 5. Comparison between our proposed baseline and Video Swin Transformer. The frame highlighted in green shows the prediction of Video Swin Transformer. The frame highlighted in blue shows the baseline's prediction. The groundtruth PNR is in **bold and underlining**.

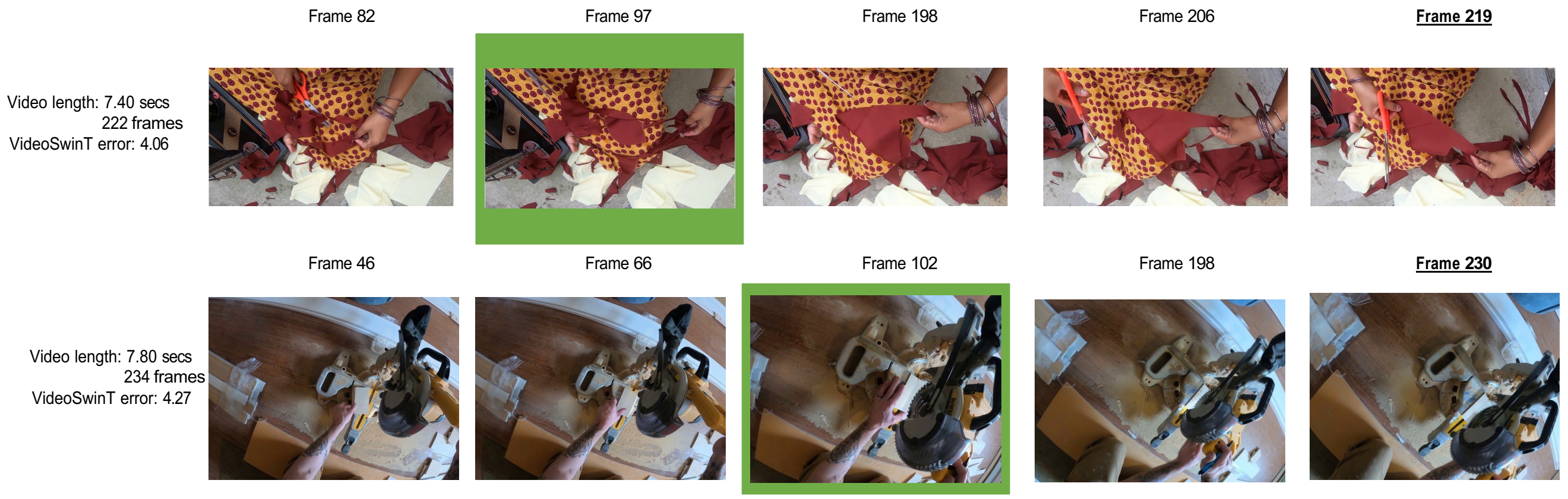

Figure 6. Failure cases of our Video Swin Transformer. The frame highlighted in green shows the prediction of Video Swin Transformer. The groundtruth PNR is in **bold and underlining**.